# Generative Adversarial Networks for Labeled Acceleration Data Augmentation for Structural Damage Detection

Furkan Luleci[1]; F. Necati Catbas[2*], Ph.D., P.E.; Onur Avci[3], Ph.D., P.E.

[1]Doctoral Student, Department of Civil, Environmental, and Construction Engineering, University of Central Florida, Orlando, FL, 32816, USA (Email: furkanluleci@Knights.ucf.edu)

[2]Professor, Department of Civil, Environmental, and Construction Engineering, University of Central Florida, Orlando, FL, 32816, USA (Email: catbas@ucf.edu)

[3]Assistant Professor, Wadsworth Department of Civil and Environmental Engineering, West Virginia University, 1306 Evansdale Drive, Morgantown, WV, 26506, USA (Email: onur.avci@mail.wvu.edu)

**Abstract:** There has been a major advance in the field of Data Science in the last few decades, and these have been utilized for different engineering disciplines and applications. Artificial Intelligence (AI), Machine Learning (ML) and Deep Learning (DL) algorithms have been utilized for civil Structural Health Monitoring (SHM) especially for damage detection applications using sensor data. Although ML and DL methods show superior learning skills for complex data structures, they require plenty of data for training. However, in SHM, data collection from civil structures can be expensive and time taking; particularly getting useful data (damage associated data) can be challenging. The objective of this study is to address the data scarcity problem for damage detection applications. This paper employs 1-D Wasserstein Deep Convolutional Generative Adversarial Networks using Gradient Penalty (1-D WDCGAN-GP) for synthetic labelled acceleration data generation. Then, the generated data is augmented with varying ratios for the training dataset of a 1-D Deep Convolutional Neural Network (1-D DCNN) for damage detection application. The damage detection results show that the 1-D WDCGAN-GP can be successfully utilized to tackle data scarcity in vibration-based damage detection applications of civil structures.

## 1) Introduction

Man-made or environmental stressors tend to decrease the remaining useful lives of civil structures. As the ageing civil infrastructures are getting more vulnerable to such impacts, more comprehensive assessment and effective health management plans are needed to improve the life cycle of structures. The typical workflow to monitor and assess an existing civil structure starts with collecting operational data with sensors such as accelerometers, strain gauges, potentiometers, fiber optic sensors or load cells. In the following step, the data is pre-processed and analyzed to perform damage identification based on the changes in the structural parameters (stiffness, mass, damping, etc.) or in the raw data to identify structural defects (crack, delamination, corrosion, bolt-loosening, spalling, etc.) One common practice of diagnosing damage in SHM is based on dynamic signature monitored via accelerometers since this has certain advantages over other methods [1].

Applications of damage detection can be categorized into local methods and global methods. Local methods include Non-Destructive Testing (NDT) and some camera sensing techniques like Infrared Thermography (IR), and Digital Image Correlation (DIC). Global methods (vibration-based) include the analysis of collected data parametrically (using a physical model like FEA software or a non-physical model like system identification methods) or nonparametrically (extracting meaningful features from the collected raw data; in other words, damage detection without using structural parameters) [2]. Additionally, via advanced Computer Vision methods, damage detection can be performed at local and global levels [3].

Numerous studies have been reported for structural damage detection of civil structures. The study presented in [4] demonstrated that the boundary conditions of a structure can be determined by looking at its deflection profile which the authors obtained using test data from a laboratory grid structure. Another study detected the presence and quantified the damage by extracting the modal properties via the Natural eXcitation Technique (NeXT) and Eigenvalue Realization Algorithm (ERA) from the collected vibration data of transmission towers [5]. The authors in [6] utilized Auto-Regressive Moving Average (ARMA) and then Gaussian Mixture Model (GMM) to first extract the features and classify them by using the Mahalanobis distance metric on the collected vibration data. The authors successfully identified undamaged and damaged datasets. The authors in [7] utilized Auto-Regressive eXogenous output (ARX) to extract and then implemented a clustering technique to classify the undamaged and damaged cases in the collected vibration data from a grid structure. The authors in [8] focused on the vibration datasets to determine outliers by using a Genetic Algorithm (GA) based on clustering on decision boundary analysis (GADBA). One of the common details in the above-mentioned parametric and nonparametric vibration-based damage detection studies is that the amount of data is not prominent in the success of the used methods. However, these methods significantly rely on manual work and they do not apply to every dataset since they have limited generalization capability. On the other hand, ML and DL methods, in particular, a well-built and trained DL model can achieve superior accuracy on complex data structures.

The ML and DL methods require a substantial amount of data samples to train the algorithms, especially DL algorithms yield outstanding results on more and more data [9]. With superior performance on feature extraction, classification, regression, and clustering, ML and DL methods have been widely accepted in civil SHM for both parametric and nonparametric vibration-based damage detection. It is observed that Artificial Neural Networks (ANN) is the most used one in ensembled other algorithms and Support Vector Machine (SVM) comes in second for SHM applications. Some of the introduced studies are [10–16]. When using an ML model for nonparametric damage detection purposes, features from the raw data need to be extracted by using some of the computational tools such as Principal Component Analysis (PCA) or Autoregressive (AR) models, which may cause computational complexity along with other limitations [2]. On the other hand, the DL algorithms can make highly accurate predictions by learning the features directly in the raw data without having to need any extracting tools. In other words, with a correctly built model and the right training process, DL algorithms can show superior performance over ML methods. In the civil SHM area, a few unsupervised DL methods, which are mostly Autoencoders [17–19] and some supervised DL methods which are mostly Convolutional Neural Networks [20–23] are presented. Yuan et al explored 2D CNN, 1D CNN, and FNN in the classification of the damage states of earthquake ground motions, where they found that the 1D CNN model also achieved good performance accounting for the computing efficiency and prediction accuracy [24,25].

**1.2) Objective and Scope of this work**

Data collection in SHM of civil structures can be time taking, challenging, and expensive. Requesting permission from authorities to install expensive and laborious SHM systems, requesting traffic closures, etc. and needing skilled experts in the field are just some of the mentioned exhaustive tasks. Considering that only a very few civil structures have permanent SHM systems in the world, it is hard to predict the damage state of the remaining structures that do not have SHM data or very limited SHM data. As a result of these challenges, the opportunity to find useful data (damage associated data, *labelled data*) for the civil structure is rare. This creates a class imbalance complication to implement vibration-based methods with AI algorithms for structural assessment purposes. The effect of class imbalance on the classification performance of the DL model is detrimental and the impact increases when the ratio of imbalance in the dataset increases. One solution can be oversampling the already existing dataset: in other words, increasing the amount of sampled data with exact copies. Yet, in this method, the AI model does not learn the variation of the damage-associated data, but only utilizes the provided exact copies of the input, which can cause the model to overfit the training dataset. Another solution can be using random transformation methods to augment the data which requires manual work before applying it to data. Since each dataset has its own characteristics and there are diverse amounts of data, not every transformation applies to every dataset. Additionally, random transformations unintentionally change the distribution of the data [26]. Another solution can be building a Finite Element (FE) model to produce displacement, stress, or acceleration data to tackle the data scarcity by analyzing the structure under similar damage scenarios simulated in the FE model. However, this method is not reliable compared to collecting actual data from a real structure as it is quite difficult to simulate the damage cases comprehensively in an FE model [27]. These challenges can yield inaccurate results for DL models when used for vibration-based damage detection. The recent novel AI methods such as Generative Adversarial Networks (GAN) have proved state-of-the-art results for generating data samples and have superior skills for learning the data domain.

While the data scarcity hinders the utilization of the ML and DL algorithms, this study presents using 1-D GAN for data augmentation for structural damage detection applications. The GAN is utilized to generate data samples to train the 1-D Deep Convolutional Neural Networks (1-D DCNN) for performing nonparametrically (directly on raw data) vibration-based damage detection on a laboratory structure. The scope of the presented work is a good fit for representing realistic scenarios that could be encountered during the operations of civil structures. For instance, when the existing dataset is very scarce to run a DL model to perform damage detection, GAN can be employed to enhance/augment the training datasets with synthetic samples of data. The introduced methodology in this paper may open new doors for more integration of ML and DL algorithms in damage detection applications of civil structures.

The objective of this study is to investigate the integrated use of 1-D GAN and 1-D DCNN to address the data scarcity problem for nonparametric vibration-based structural damage detection (level-I damage diagnostics). Note that since this study considers a nonparametric damage detection problem, it only considers and operates on time-domain properties. First, a state-of-the-art GAN variant, 1-D Wasserstein Deep Convolutional Generative Adversarial Networks using Gradient Penalty (1-D WDCGAN-GP) is employed for synthetic labelled acceleration data generation to augment the training data with varying ratios. Subsequently, the 1-D DCNN model is trained with the synthetically augmented data and then tested on the unseen raw acceleration data to demonstrate the performance of the WDCGAN-GP. Fig. 1 represents the objective of the study with a schematic diagram. It is a well-known fact that the imbalanced data classes are detrimental to the performance of DL models which lowers their prediction scores. This study aims to solve the imbalanced data classes problem due to the data scarcity issue in the SHM field. For that, the amount of data samples in the damaged class is reduced with five different ratios (five different scenarios). Then, 1-D WDCGAN-GP is used to augment the damaged class on five different levels. Subsequently, a 1-D DCNN model is trained with a naturally balanced dataset and tested on the unseen dataset. Then, the same model is trained with the synthetically augmented dataset for five different scenarios (different augmentation ratios/levels in each scenario) and tested on the same unseen dataset. It is expected that the damage detection with the 1-D DCNN model trained with the augmented dataset will give a very similar prediction performance as the damage detection with a 1-D DCNN model trained with the naturally balanced dataset.

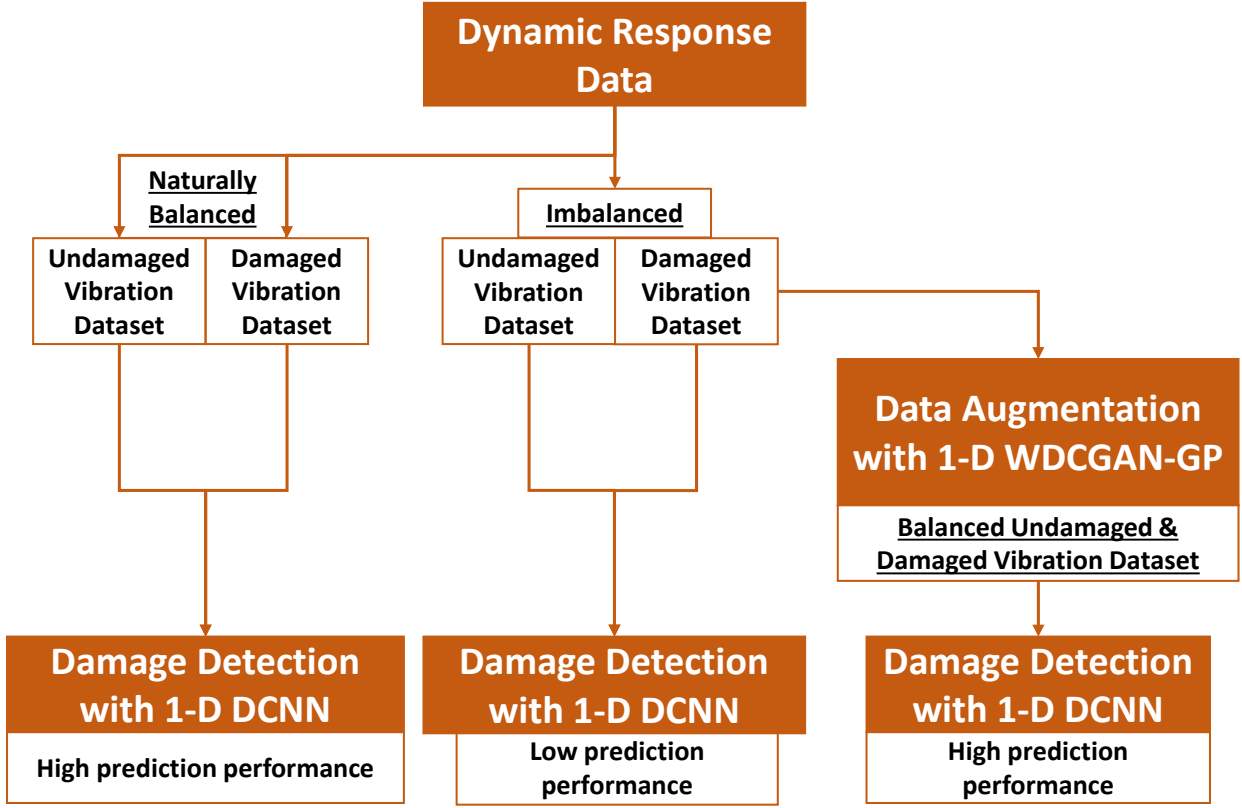

**Fig. 1.** The objective of the study

**1.3) Existing work on Generative Adversarial Networks**

The authors in [28] proposed an unsupervised DL network that contains two distinct networks: a generative network, $G_\theta$, that captures the given randomly distributed data, $z$, and aims to maximize the probability of the output it generates, $G_\theta(z)$, very alike to the training dataset, $x$, in other words, the generator tries to minimize the probability that the prediction of discriminator on generator's outputs is not real. The discriminative network, $D_\varphi$, aims to maximize the likeliness of the output it gets from the generator, $G_\theta(z)$, is not real and from the training data, $x$, is real. Thus, it is thought of as a two-player game where each player is trying to trick the other. The fundamental formulation of the loss function of GAN is shown in Equation 1.

$$\min_\theta \max_\varphi V(G_\theta, D_\varphi) = \mathrm{E}_{x \sim p_{data}(x)}[log D_\varphi(x)] + \mathrm{E}_{z \sim p_z(z)}[log(1 - D_\varphi(G_\theta(z)))] \qquad (1)$$

The GAN is successfully used on image-based applications and then adopted in several other applications. At the time when the GAN was first introduced, the training procedure of GAN was not a trivial task due to several difficulties as explained in the following sentences. For example, reaching a convergence due to the difficulty of finding a unique solution to Nash equilibrium where the model is searching for a balance point between the two sides in the optimization, instead of finding a minimum one. Consequently, GAN experiences large oscillations in loss values which leads to an unstable training process and model performance. Occasionally, GAN suffers from mode collapse, in another word, the generator learns only some specific features in the given training data where the discriminator can be deceived. Thus, the GAN keeps producing the same data samples to trick the discriminator which reduces the creativity of the model and the diversity of the outputs. GAN also suffer from the discriminator being powerful over the generator.

As a result, training the generator could fail due to vanishing gradients, hence, the discriminator does not provide enough information for the generator to learn [29,30]. Although the mentioned drawbacks may make it harder to train GAN, the following studies improved the GAN model significantly, which is discussed in the next paragraph. Additionally, tips, namely "hacks" are introduced in the DL community to alleviate these drawbacks and some of which are used in the presented work herein.

The researchers in [31] introduced the use of convolutions in GAN (and called it DCGAN) after their successful adoption in computer vision applications. The authors noted that convolutions helped GAN to learn significantly better. However, to address the challenges in the training of GAN, the authors [32] introduced GAN that uses Wasserstein distance as a loss function (Wasserstein Generative Adversarial Networks - WGAN) and demonstrated that it improves the training remarkably. In WGAN, the authors used "critic" over discriminator where instead of scoring the likelihood of the generated data being real or fake, it scores the output of the generator as how close to the real or fake on a given image. Essentially, WGAN looks for a minimization of the distance between the produced and the training dataset distributions in the range of a real or fake distance. GAN benefited greatly from using the Wasserstein metric. It provided more stable training; the results are less sensitive to the parameters and model architecture, and the loss function is more meaningful which directly relates to the quality of generated images. WGAN uses weight clipping to enforce the Lipschitz constraint on the critic to compute the Wasserstein distance; however, it causes the model to lower its learning capacity. It is also very sensitive to the selected parameters. If large weight clipping is used, it increases the computation time drastically; if the clipping is small, it could easily lead to vanishing gradients. To tackle the weight clipping problem, the authors [33] proposed a penalization of the gradient during the training of the critic instead of using weight clipping. The authors showed that the method they introduced yields better performance than WGAN and resulted in more stable training and the authors named presented the model as **WGAN-GP (Wasserstein Generative Adversarial Networks using Gradient Penalty)**. The formulation of WGAN-GP used in this study is given in Equations 2 and 3, where $z$ is the noise domain, $y$ is the target domain (in this study it is the damaged dynamic response domain), $\tilde{y}$ is the generated synthetic domain, $\lambda_{GP}$ is the adjustable weight parameter for the gradient penalty term.

$$Adversarial\ Critic\ Loss = \mathcal{L}_{WGAN-GP}^{Critic}(Generator, Critic, z, y, \tilde{y}) =$$

$$\mathbb{E}_{y \sim p(y)}[Critic(y)] - \mathbb{E}_{z \sim p(z)}[Critic(Generator(z))] + \lambda_{GP}\mathbb{E}_{\tilde{y} \sim p(\tilde{y})}\left[\left(\left\|\nabla_{\tilde{y}}Critic(\tilde{y})\right\|_2 - 1\right)^2\right] \quad (2)$$

$$Adversarial\ Generator\ Loss = \mathcal{L}_{WGAN-GP}^{Generator}(Generator, Critic, x, y) =$$

$$-\mathbb{E}_{z \sim p(z)}[Critic(Generator(z))] \quad (3)$$

The computer Vision field largely benefited from using GAN on various image applications which are 2-D data. Additionally, some studies in different disciplines were successful at using GAN for 1-D data generation and data reconstruction for various purposes [34–39]. Several GAN-based 1-D data generation, reconstruction, and ML classifier-related studies are presented in the field of non-civil SHM [40–43]. In the civil SHM field, few studies are presented for missing data reconstruction using 1-D GAN [44–46]. While those studies only explore the missing data reconstruction, more recently, the authors in [47] attempted to implement dynamic response generation using WGAN-GP. The authors trained their DCNN model on the real undamaged and damaged data classes and tested the model on the real data undamaged and synthetically generated damaged classes. The results showed that DCNN was able to identify the synthetic damaged class as a real damaged class. That study demonstrated that WGAN-GP can be employed for data augmentation for imbalanced classes in future work.

On the other hand, the presented study herein conducts a WGAN-GP-based data augmentation on a training dataset and uses DCNN-based damage detection on both augmented and unaugmented datasets. In other words, this study employs WGAN-GP to augment the training dataset with varying augmentation ratios (5 different scenarios). Then performs damage detection using DCNN on the augmented training datasets and unaugmented dataset (naturally balanced dataset) to evaluate the performance and applicability of the data augmentation process for damage detection.

**2) Methodology**

In the proposed work, 1-D WGAN-GP built on convolutions is used to generate synthetic datasets, which is named "1-D WDCGAN-GP" in short by the authors for One Dimensional Wasserstein Deep Convolutional Generative Adversarial Networks using Gradient Penalty. Then, the synthetic datasets are used to augment the training dataset of 1-D DCNN with varying ratios (5 scenarios with different augmentation ratios). Subsequently, the 1-D DCNN model is trained for each of these scenarios and additionally trained with a reference scenario, which is only consisted of real datasets. Then, the model is tested on the real dataset. Thereafter, the 1-DCNN model is evaluated for its performance considering the test results on the training dataset with varying ratios of augmentation (5 scenarios) and reference scenarios. For simplicity, in the remaining of the paper, 1-D WDCGAN-GP and 1-D DCNN are referred to as $M_1$ and $M_2$, respectively. The rest of the paper is presented in the following order: the methodology of the study in section 2, the workflow of the 1-D WDCGAN-GP in section 3, and the workflow of the 1-D DCNN in section 4. Summary and conclusions are presented at the end.

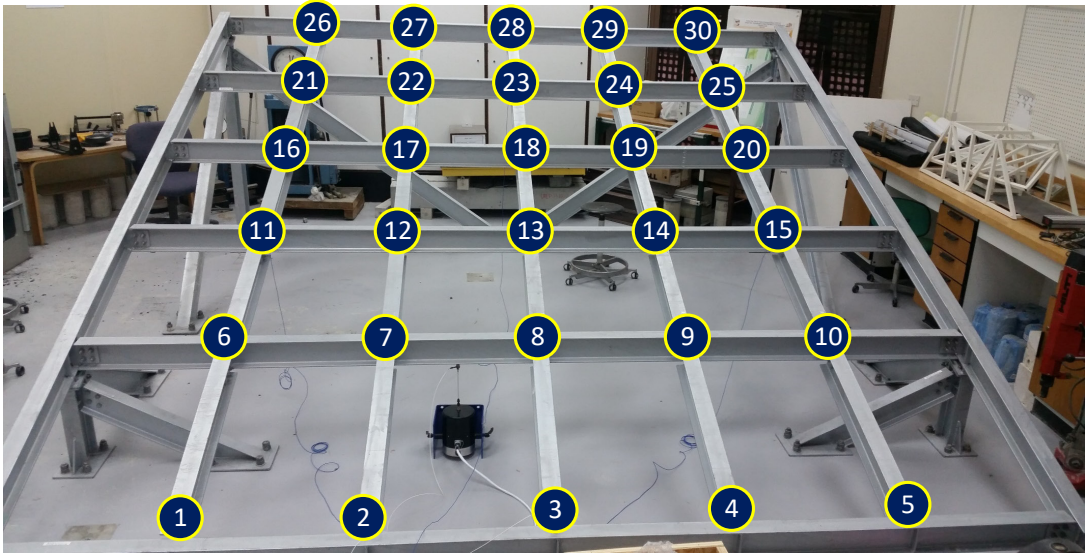

**Fig. 2.** Steel Frame Grand Simulator Structure [22]

The vibration dataset used in this work is obtained from a study by [22] on a steel laboratory frame (Fig. 1) where they installed 30 accelerometers at 30 joints. A modal shaker random excitation, as white noise, is applied to the structure and consequently, 1 undamaged and 30 damaged scenarios are created at each joint separately by loosening the bolts at the steel connections between filler beams and girders. Then, at a 1024 Hz of sampling rate, the authors collected 256 seconds of vibration data with a total sample of 256 x 1024 = 262,144 from each accelerometer channel dedicated to a joint (Fig. 1). The notation used in this paper for acceleration tensors follows $n[a_{c,s}^t]$ where $c$ represents the condition such as $u$ refers to the data collected in an undamaged scenario, $d$ refers to the damaged scenario, and $s$ refers to synthetic if the tensor is generated by the $M_1$. Lastly, $t$ refers to the seconds that the tensor contains, and $n$ refers to the number of tensors. The *n* is featured to 1 if there is no number written. In this study, the *t* is only used as 1 and 256 respectively where 1 refers to "1-second" (batched from 256 seconds) and 256 refers to "256-seconds" (entire raw signal). Additionally, in this study, only Joint 1's datasets are utilized. Lastly, the PC used in this study has the following specs: 16 GB RAM DD4 2933 MHz and NVIDIA GeForce RTX 3070 8GB GDDR6 graphic card.

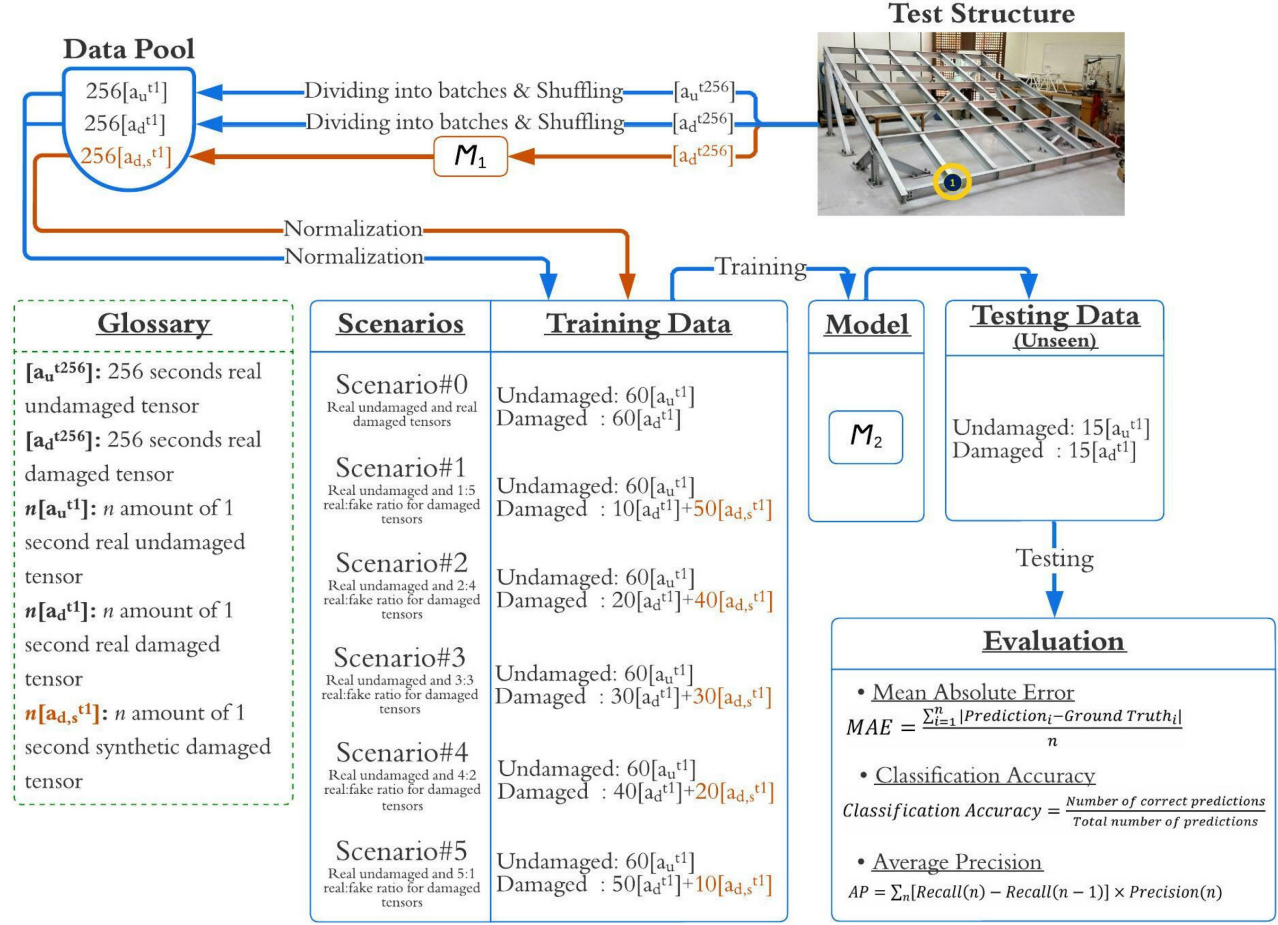

**Fig. 3.** The framework

The workflow of this study is presented in Fig. 2. First, the tensor $[a_d^{t256}]$, 256 seconds of real damaged tensor, is input in the $M_1$ to generate the synthetic datasets of $256[a_{d,s}^{t1}]$, which is 256 amounts of 1 second synthetic damaged tensors. The $[a_u^{t256}]$ (256 seconds real undamaged tensor) and the $[a_d^{t256}]$ (256 seconds real damaged tensor) are batched into 256 amounts of 1-second tensors, which created $256[a_u^{t1}]$ and $256[a_d^{t1}]$. Then, they are randomly shuffled and placed in the data pool without replacement. Two reasons why batching in shuffle mode is used: (i) to prevent any bias in the model such as learning the data in the order that is provided and (ii) to converge the training faster. A similar method is also successfully implemented and described by [20]. The authors divided the input signal into frames and randomly shuffled it. Therefore, the sequence of the dataset or the signal's spectral properties or parameters of the vibration tensors is irrelevant because the purpose of the study is to show that GAN can tackle the data scarcity for nonparametric (directly on raw data) damage detection for civil structures. After the normalization and then grouping in the training data ($M_1$ outputs are not normalized which is explained in the next section), these tensors are passed in $M_2$ for training and testing for six different scenarios of which five of them include varying ratios of data augmentation (Scenario#1-5). Note that the datasets that are used in training are not used in testing.

In Scenario#0, for benchmarking purposes with other scenarios, no synthetic data is involved, but only real tensors are used. To represent that, $60[a_u^{t1}]$ and $60[a_d^{t1}]$ are used for training which includes 1-second of real undamaged and real damaged tensors. In Scenario#1 to Scenario#5, real undamaged tensors are used for both training and testing. However, the real damaged tensors are augmented with varying ratios of synthetic tensors for Scenario#1 to Scenario#5. For instance, in Scenario#1, while the undamaged class has $60[a_u^{t1}]$ (60 amounts of 1 second of the real undamaged tensor), the damaged class has $10[a_d^{t1}]$ (10 amounts of 1 second of the real damaged tensor) and $50[a_{d,s}^{t1}]$ (50 amounts of 1 second of the synthetic damaged tensor). The same logic can be extrapolated to other scenarios as illustrated in Fig. 2. For testing $15[a_u^{t1}]$ and $15[a_d^{t1}]$ tensors are used. The reasoning behind creating the scenarios in this way is to reflect the possible scenarios that can be encountered in SHM to perform damage detection of civil structures. Additionally, introducing different ratios of augmentation demonstrates the extent of augmentation that can be employed for data augmentation and damage detection problems. As such, it is important to demonstrate the effect of the data augmentation in the training dataset and its subsequent impact on the damage detection prediction results by the $M_2$ on the unseen data. Thus, it is decided to set 5 augmentation scenarios with varying ratios of augmentation levels and 1 reference scenario with no augmentation to evaluate the effect of data augmentation on the training dataset (having more scenarios could further validate the methodology with additional computation cost – in this study setting 5 scenarios is found sufficient). As previously mentioned, obtaining damage-associated datasets from large civil engineering structures is challenging; and to implement damage detection via a DL model on a particular structure, it is essential to have a sufficient amount of data samples for each class (undamaged or damaged classes).

Each scenario presented in Fig. 2, from Scenario#1 to Scenario#5, shows that enough amount of tensor exists in the undamaged class, but the tensor in the damaged class is very scarce for the particular damage case. This case is most of the time true for bridge structures. After further validated investigations and developments on the introduced methodology herein, the authors of this study envision that the methodology may be used for bridge structures. For instance, single-channel vibration datasets were previously collected on the first 6 spans out of 18 spans of a multi-span bridge. It is previously known that the first 5 spans contain undamaged, and the 6[th] span contains damaged features in the dataset. To diagnose the condition of the remaining 12 spans via a DL model (which might potentially house similar types of damage), it is critical to have a trained DL model on that damage-associated dataset. Yet, under given circumstances, the dataset suffers from class imbalance due to data scarcity (5 undamaged and 1 damaged dataset out of a total of 6 datasets corresponding to 6 spans). To tackle this problem, GAN may be used to generate synthetic data to augment the training dataset of the DL model. This instance is illustrated in Fig. 3 and Fig. 4. Fig. 3 illustrates the vibration-based damage detection via the DL model. Fig. 4 illustrates the GAN augmented vibration-based damage detection via the DL model. Then, the following sections explain the $M_1$ and $M_2$ workflows.

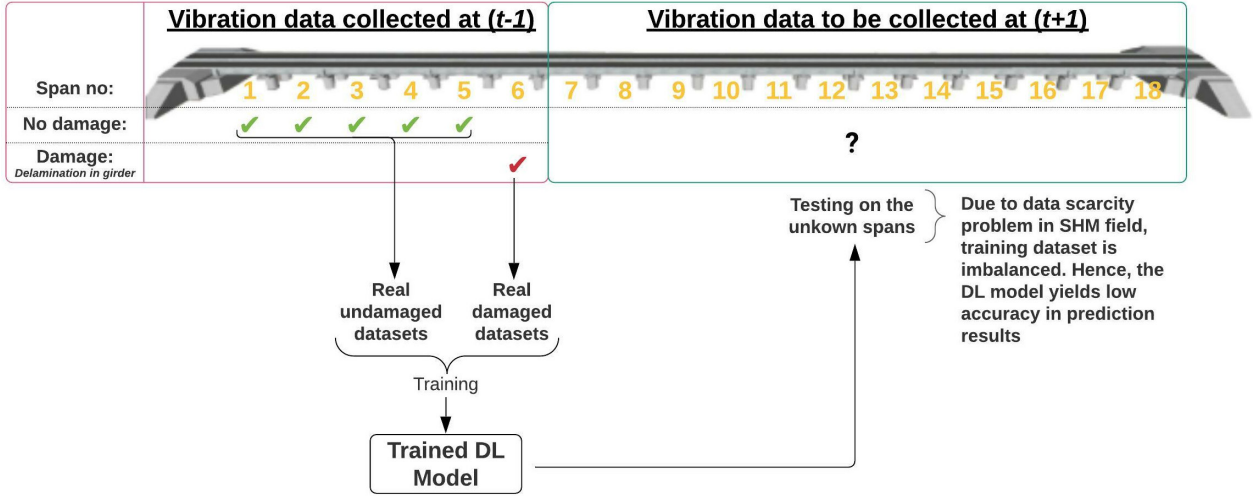

**Fig. 4.** Vibration-based damage detection via DL model

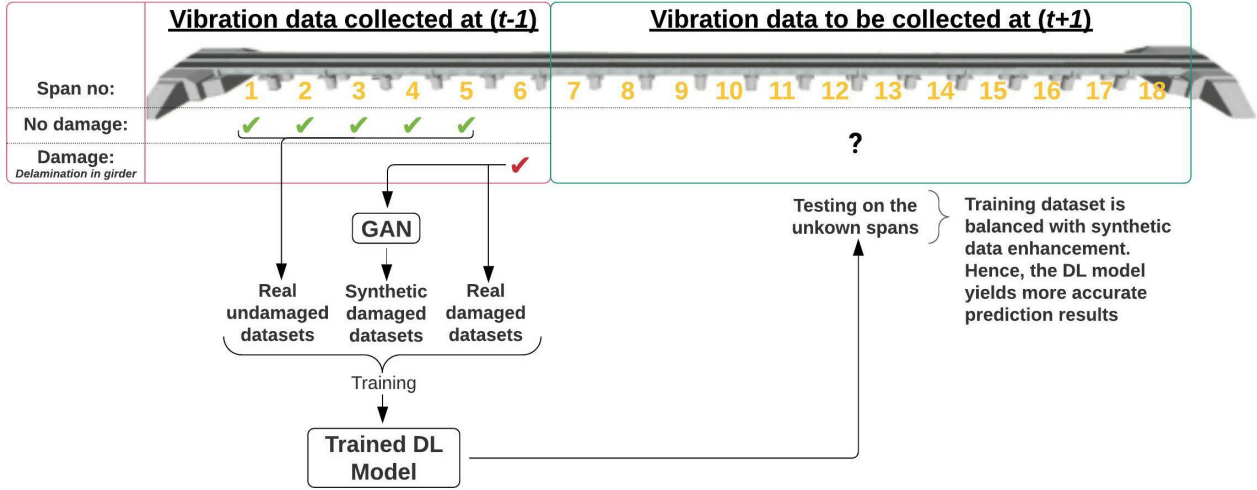

**Fig. 5.** GAN enhanced vibration-based damage detection via DL model

### 3) Workflow of 1-D WDCGAN-GP ($M_1$)

#### 3.1) $M_1$ - data preprocessing

Before the training of the DL model, the standard practice in the DL field is to normalize the inputs on the same scale. Thus, the model can learn and predict more efficiently. During the model training, normalizing helps with the weights to be at the same scale. In other words, during the forward and backpropagation through the network where the dot products of the weights are computed, the normalization assists the model to perform accurate results and decrease processing time. In case the dataset has large spikes, the normalization would help. If normalization is not used, the spikes have a significant impact on the propagation because different levels of weight calculations can negatively affect the model quality during training. Thus, normalization is carried out for the outputs of $M_1$ before using the inputs for the $M_2$.

#### 3.2) $M_1$ – architecture

As a primary step, different options of layers and parameters are used in the architecture and the one with the best performance is selected. The model architecture used is illustrated in Fig. 5 along with the filter, stride, and padding sizes used in the convolution layers. The architecture takes the randomly given noise tensor ($z$) and passes it through the five 1-D convolutions. In the generator, batch normalization and ReLU are used after every layer except the last one. At the end of the generator part, the Tanh activation function is used and consequently, $[a_{d,s}{}^{t1}]$ created. Then, $[a_d{}^{t1}]$ and $[a_{d,s}{}^{t1}]$ are passed to the critic (the "discriminator" is named in WGAN "critic" as in their original paper) to be scored by the critic as to how real or fake each passed tensor is. The critic consists of five 1-D transpose convolutions. In critic, instance normalization and Leaky ReLU are used after every layer except the last one.

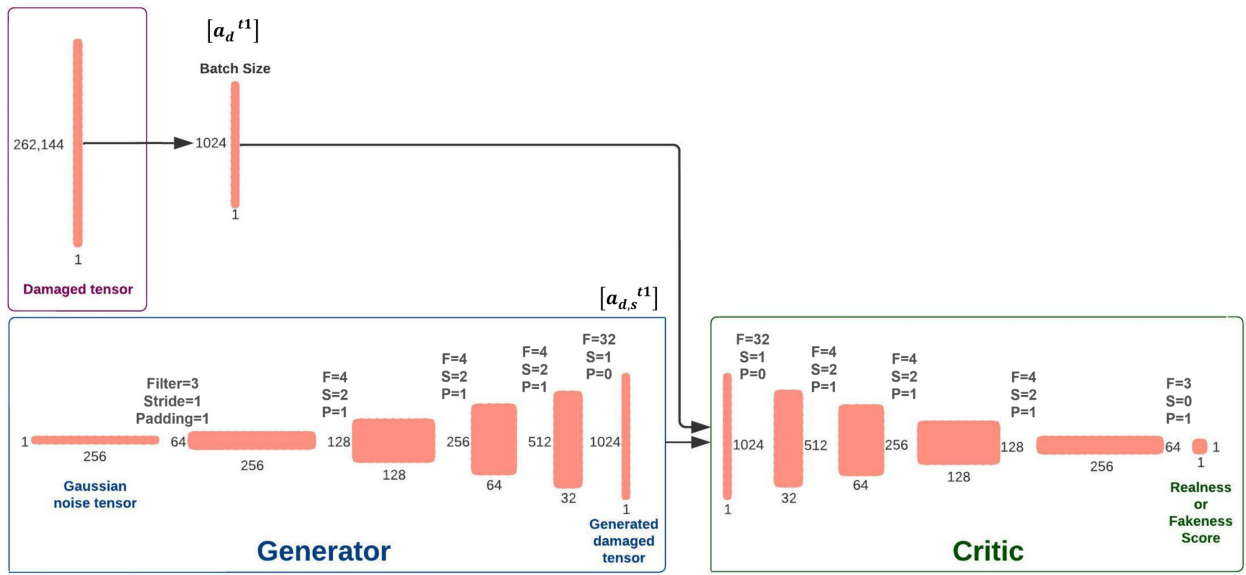

Fig. 6. 1-D WDCGAN-GP (M1) Architecture

### 3.3) $M_1$ - training and fine-tuning

Throughout the training procedure of 1-D WDCGAN-GP, few approaches (so-called hacks) have been taken for fine-tuning. After plenty of trials with different parameters in the model, using one layer of dropout with 70% in the critic is found very beneficial which avoids overfitting and reduces the capacity of the critic and helps the model to reach the Nash equilibrium. Moreover, a random Gaussian noise is added that decays over each epoch (iteration). Thus, it gives a handicap to the critic not to be superior to the generator. Consequently, the learning rate of $5 \times 10^{-6}$ for generator and $2 \times 10^{-5}$ for critic yielded the best result. The critic iterations, lambda parameter for the gradient penalty, and batch size are respectively picked as 12, 20, and 1024. The epoch number is used as 600 and took 44 hours of training. The AdamW optimizer is used in both the generator and critic for the optimization process. Lastly, the generator and critic loss functions are seemed to be converged to zero (Fig. 6) which is explained in detail in the next section.

### 3.4) $M_1$ - evaluation and interpretation

Arguably, the evaluation of GAN models is quite difficult in the DL field. Generally, the evaluation can be categorized as qualitative (manual evaluation) and quantitative where the former is based on visual, and the latter is based on numerical evaluation. The widely used form of evaluation method of GAN is a qualitative approach that is visually evaluating the outputs of the generator in comparison with the training data, and the data is mostly in the form of images. However, this approach suffers from some drawbacks such as the limited number of generated outputs that can be viewed at a time by an observer and subjectivity being introduced by observers. Additionally, implementing this approach is not as effective as doing it on 1-D signal data. Therefore, several quantitative evaluation metrics are introduced to evaluate the performance of the model with no reached consensus in the DL as to which metric is the most effective. Because this study considers only nonparametric structural damage detection, the generated tensors (signals) are evaluated only in the time domain. In a study by Borji (2018) the evaluation methods of GAN are investigated. One of the most used indicators for evaluating GAN is the Fréchet Inception Distance (FID) score [49] which is introduced as an improvement over the Inception Score (IS) which is reported to be unable to capture the distribution of real and generated output. Also, several comparative studies proved its effectiveness against other GAN evaluation metrics. Particularly, FID showed very consistent performance when compared to the manual evaluation of the GAN outputs. The formula of FID is based on a statistical formulation that is shown in Equation 2.

$$FID(x, g) = ||\mu_x - \mu_g||_2^2 + Tr\left(C_x + C_g - 2(C_x C_g)^{0.5}\right) \quad (2)$$

Where respectively the $\mu_x$ and $\mu_y$ are the means and $C_x$ and $C_g$ are the covariance matrices of real and generated signals and $Tr$ is the trace of the matrices e.g., the sum of all the diagonal elements in the matrices. The lower the FID score, the more similar the compared data.

Another quantitative evaluation metric is introduced by [50] where the performance of GAN is evaluated on images based on three aspects: *Creativity*, *Inheritance*, and *Diversity*. These aspects are very significant for evaluating the GAN as they are expected to add creativity and diversity in the outputs as well as to keep the inheritance of the real input [51]. Among the three, the Inheritance aspect is generally used for images where it shows how the generated images retain the key features of the real images such as texture. Therefore, it is not used in this presented study. The Creativity aspect indicates to what extent the generated outputs are not the exact ones of the real outputs (or dissimilar to each other) and the Diversity aspect indicates to what extent the generated outputs are similar to each other. The Creativity and Diversity computations are carried out by using the Structural Similarity Index Measure (SSIM) [52] which was first used for image quality assessments by using similarity between the pixels of two images. If the SSIM of two images is 1, they are exactly the same; if the SSIM is 0, then they are entirely different images. The authors of the paper [50] defined a threshold value of 0.8 to calculate the creativity index; for instance, if the SSIM of generated and real data are higher than 0.8, they are concluded as duplicates. No threshold is used for the diversity index, but SSIM is employed between the generated datasets to compute the index. Using SSIM as an evaluation indicator is also used in [39] for evaluating the generated signals by GAN. There is no defined value in the literature on what SSIM values or thresholds should be used for creativity and diversity approaches for the evaluation of GAN. Intuitively, calculating the SSIM of the generated outputs to real inputs indicates how creative the outputs are from the inputs. Calculating the SSIM of the generated outputs to each other in the generated output dataset gives how diverse the outputs are from each other in the range of 0 between 1. The Creativity and Diversity indices are not directly used in this study, but SSIM computation is carried out to evaluate the extent of creativity and diversity of the generated outputs. As a result, in this study, the creative approach is investigated by computing SSIM between the generated and real tensors; the diversity approach is investigated by computing SSIM between the generated tensors within the generated dataset from the $M_1$. But using SSIM may not be a valid approach for evaluating 1D outputs of GAN and further investigations are needed since using them is only evaluated by one study as mentioned above. The SSIM equation is:

$$SSIM(x, g) = \frac{(2\mu_x\mu_g + c_1)(2\sigma_{xg} + c_2)}{(\mu_x^2 + \mu_g^2 + c_1)(\sigma_x^2 + \sigma_g^2 + c_2)} \quad (3)$$

In Equation (3) $\mu_x$ and $\mu_g$ are the means, $\sigma_x$ and $\sigma_g$ are the standard deviations, and $\sigma_{xg}$ is the covariance of real data ($x$) and generated data ($g$). The $c_1$ and $c_2$ are the constants which are multiplication of $k_1$ and L; and $k_2$ and L respectively, to stabilize the division with a weak denominator. L is the dynamic range of the signal and $k_1$ and $k_2$ are the constants which are picked in this study as $1\times10^{-2}$ and $3\times10^{-2}$.

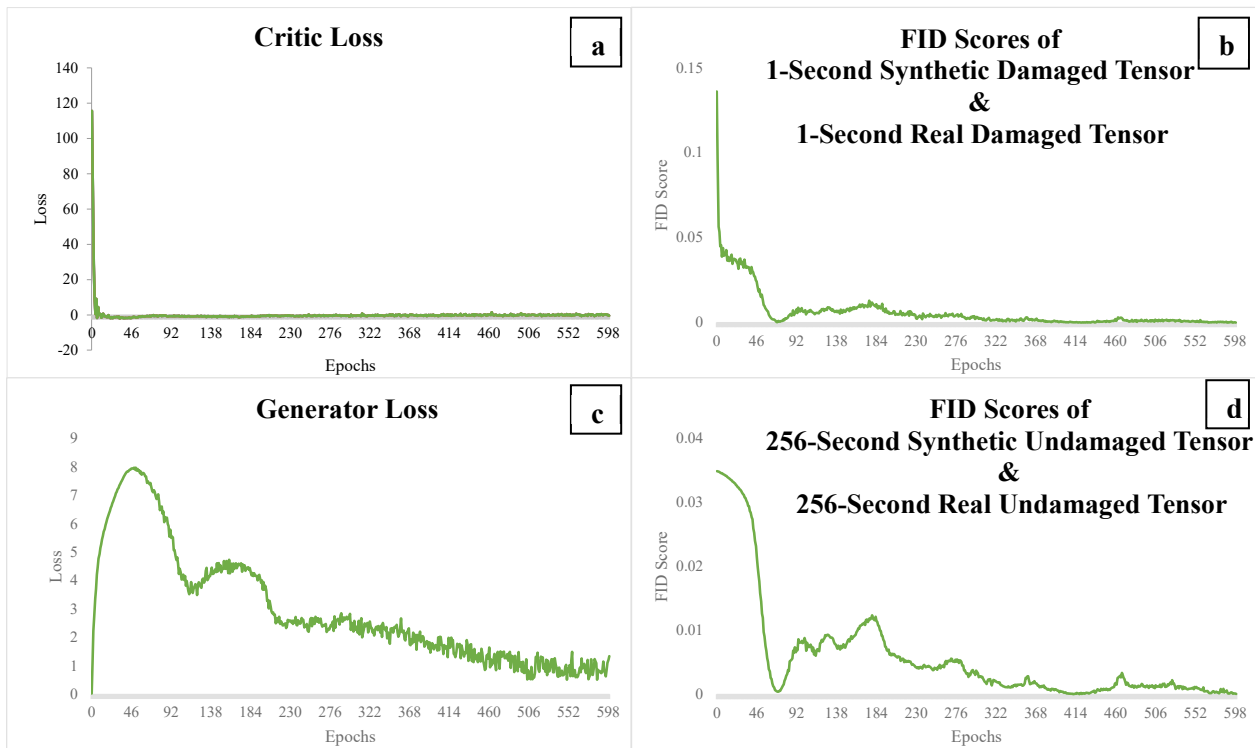

**Fig. 7.** Training plots of 1-D WDCGAN-GP ($M_1$)

During the training, the critic loss, the generator loss, and FID scores are monitored to observe the performance of the model (Fig. 6). It is noted that the critic loss is converged to zero. The generator loss on the other hand heuristically starts from near zero and then increases since the critic has more knowledge of the real data domain and can easily guess the generated outputs. After some epochs, the generator starts learning the gradients and generates more similar datasets. Hence, the generator loss seems to be returning to its first loss values which are expected for WDCGAN-GP. The FID scores are monitored by computing the tensors between $[a_{d,s}{}^{t1}]$ and $[a_d{}^{t1}]$ and they appear to be converging to zero. In other words, they are getting to look like each other (Fig. 6b and 6d). It is also important to note that the generated outputs, $[a_{d,s}{}^{t1}]$, are not only becoming similar to batched real data, $[a_d{}^{t1}]$, but also to the entire raw signal data, $[a_d{}^{t256}]$. The FID between $[a_{d,s}{}^{t256}]$ and $[a_d{}^{t256}]$ is calculated during the training. It is found that the FID scores of $[a_{d,s}{}^{t1}]$, and $[a_d{}^{t1}]$ and the FID scores of $[a_{d,s}{}^{t256}]$ and $[a_d{}^{t256}]$ are both converging to zero with the same trend over the epochs. This reveals that batch sampling from the raw data, $[a_d{}^{t256}]$, in shuffle mode, does not produce different results since it suggests that the 262,144 sized raw datasets have repetitive features in a particular time length. Hence, it validates using the batch samples of 1-second tensors (1024 samples in each tensor) in this study. Note that since the aim of this study is nonparametric damage detection, which is not based on any parameters but directly on raw data, the order of the samples in the batch sampled (1-second) tensors is irrelevant. Batch sampling in shuffle mode also helps the training of the model in a way that it converges faster, preventing bias, and preventing learning the order of the data. Therefore, for simplicity, in the rest of the study, only the calculations between synthetic damaged and real damaged datasets ($[a_{d,s}{}^{t1}]$ and $[a_d{}^{t1}]$) are considered. Furthermore, in Fig. 6, it is observed that the FID score of $[a_{d,s}{}^{t1}]$ between $[a_d{}^{t1}]$ are started from 0.136 and decreased to $1.5\times10^{-5}$, which is a reduction of 9060 times. This large reduction shows that the model learned the dataset thoroughly and can generate very similar datasets. The FID scores can depend on various factors (model architecture and hyperparameters and used dataset) as to how much it reduced, or where it started and ended. To better comprehend the FID values, in one study by [53], the authors compared their GAN model to others on MNIST data (image dataset of handwritten numbers) and the FID score was decreased by 6 times during the training. In another study, the resulting decrease in FID score was found around 85 [54]. Although a decrement of 9060 times can be looking like a success, it might also be an indication of overfitting which will be investigated in the following paragraph.

After the $M_1$ is trained, the FID scores are computed between the real and generated tensors, $[a_d{}^{t1}]$ and $[a_{d,s}{}^{t1}]$. Then, the Probability Density Function (PDF) is plotted of the calculated FID scores (Fig. 5). It is seen that the FID scores are dense around the value of $7 \times 10^{-4}$. The purpose of visualizing the PDF is to comprehend the variance of FID values. Although during the training it went down to $1.5 \times 10^{-5}$, the variance of the FID scores could have been extended to a value of 0.10 as well, which in this case it could be concluded that the model did not learn. However, as seen in Fig. 7, the variance is considerably low along with very low FID values, 194 times lower than the starting point of 0.1359.

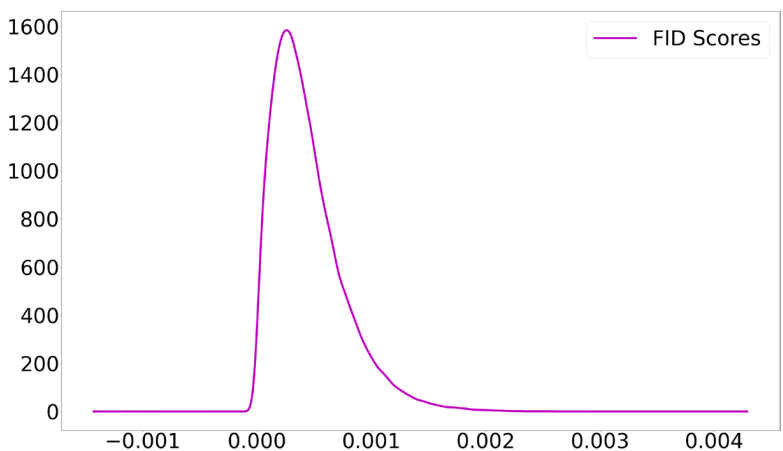

**Fig. 8.** Probability Density Function plot of FID Scores

Next, the creativity and diversity approaches are investigated. For that, the SSIM between the $[a_{d,s}{}^{t1}]$ and $[a_d{}^{t1}]$, and SSIM of the tensors of $[a_{d,s}{}^{t1}]$ are computed respectively. The PDFs of the creativity and diversity results are plotted in Fig. 8. The SSIM values do not go beyond the 0.8 threshold value, and they are dense around the value of 0.3 which may be concluded that the generated tensors are not the exact copies of the real tensors. Yet, further validations are needed to support this claim since SSIM was used only once in the literature for 1-D data generation [39]. Thus, $M_1$ can generate creative outputs. The creativity measure also determines the overfitting of the model; considering that the resulting value is small, it can be concluded that the model is not overfitted in the training dataset since no exact copies exist. The diversity values are dense around the value of 0.36 and it can be stated that the generated tensors are not alike to each other. This can mean that diversity exists within the generated tensors. Based on the creativity and diversity aspects of the generated dataset, none of the generated and real tensors is the exact copies of each other and none of the generated tensors is exact copies of each other.

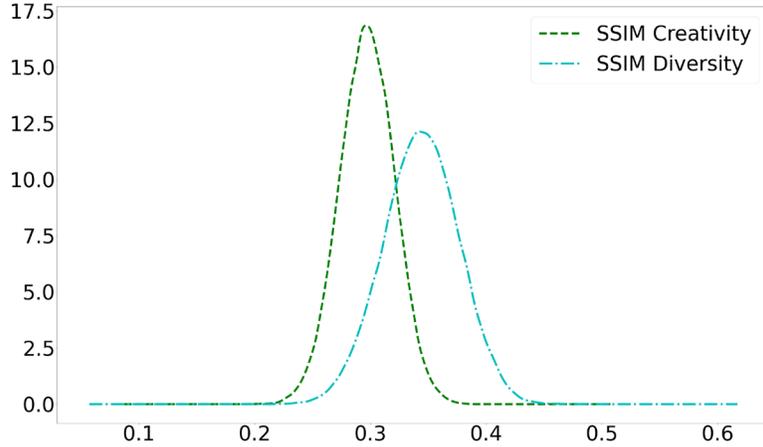

**Fig. 9.** Probability Density Function plots of Creativity and Diversity by using SSIM

Two pairs of vibration tensors two from real and two from the generated dataset are boxes plotted in Fig. 9 to visualize and compare the statistical meanings. The plotted tensors are picked as one pair has the lowest FID, and one pair has the lowest FID score. The boundaries of boxes and the size of whiskers are looking alike along with the mean and median values. The box plots show that the statistical meanings and distributions of the generated and real data pairs are very similar to each other.

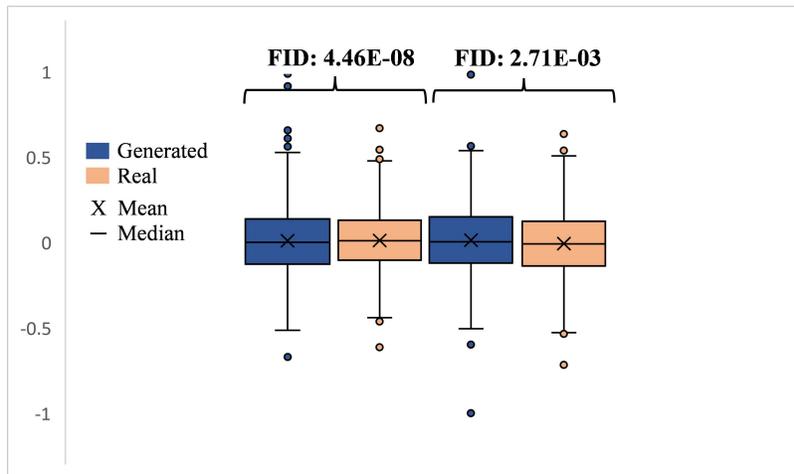

**Fig. 10.** Box plots of the tensors with lowest and highest FID values

Lastly, qualitative evaluation is implemented on the real input and generated outputs of $M_1$. As previously mentioned, qualitative evaluation is the most preferred method for image data, 2-D data, yet it has drawbacks for evaluating 1-D data. The same vibration tensors that were used in Fig. 9 are plotted in Fig. 10. Although it is difficult to determine the similarity between the generated and the real data as in image-based applications, the plotted tensor pairs in Fig. 10 reveal good consistency throughout the sample length for the lowest and highest FID scores. Also, while the generated tensors have some slight spikes, the 1-D DCNN ($M_2$) model's predictions were successful after the data augmentation, which is explained in section 4.4.

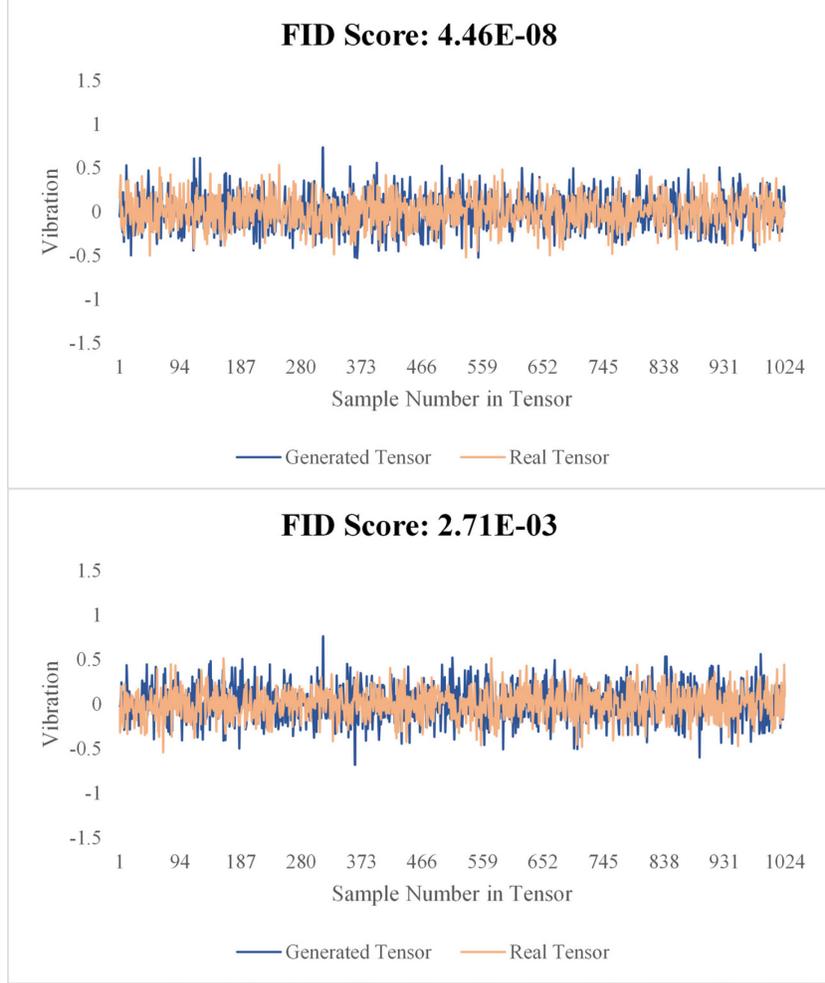

**Fig. 11.** Tensor pairs with lowest and highest FID values

**4) Workflow of 1-D DCNN ($M_2$)**

**4.1) $M_2$ - data processing**

Before feeding the tensors in the $M_2$ (1-D DCNN), the tensors are normalized in the range of -1 and +1. The generated tensors from $M_1$, $[a_{d,s}{}^{t1}]$, and 1-second tensors from $[a_d{}^{t256}]$ and $[a_u{}^{t256}]$ are randomly extracted into the data pool. Then, the datasets are arranged to be used for each different augmentation scenario as shown in Fig. 2.

**4.2) $M_2$ - architecture**

The same critic architecture used in $M_1$ is utilized for $M_2$ but only having a sigmoid function at the end of $M_2$ in order to produce a prediction score for each tensor (the critic network in $M_1$ had no activation function at the end of the last layer and only realness or fakeness scores were processed based on the Wasserstein distance). The sigmoid function produces prediction scores in the range of 0 to 1 where 0 denotes undamaged and 1 denotes damaged tensor. Also, unlike in the critic network of $M_1$, no dropout is used in $M_2$ since it is not considered necessary for a simple detection process. Lastly, the test or validation loss indicators are not plotted in this study since the prediction results for the unseen data demonstrated that the model can successfully classify the unseen tensors. This is presented in section 4.4.

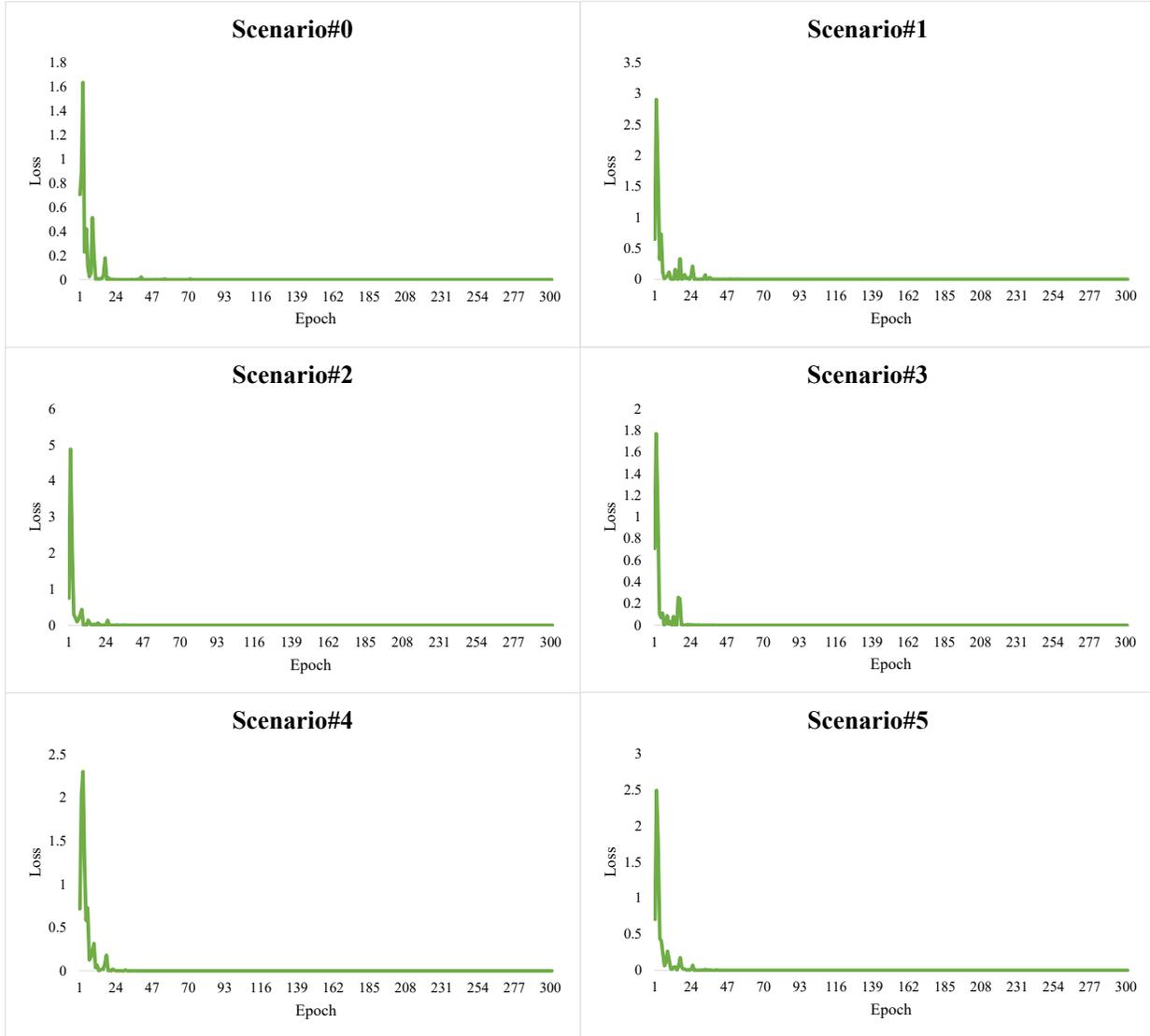

**Fig. 12.** Training plots of $M_2$ (1-D DCNN) for each scenario

### 4.3. $M_2$ – training and fine-tuning

After several trials of using different hyperparameters in the $M_2$, the learning rate for Scenario#0, Scenario#1, and Scenario#2 is chosen as $8 \times 10^{-4}$ and for the rest of the scenarios, it is $3.5 \times 10^{-3}$. The batch size and number of the epoch are picked as 30 and 300, respectively. Also, Cross-Entropy is used as the loss function which was not used in $M_1$. Finally, the training loss functions are plotted in Fig. 11 which are seemed to be converged at zero for all scenarios, in other words, the model learned the training dataset. Then, the model is tested on the unseen tensors for each scenario and yields accurate prediction results, which are explained in detail in the following section.

### 4.4) $M_2$ - evaluation and interpretation

Regardless of how the model learned the training dataset successfully, the testing phase determines the performance of the model (the testing dataset contains instances that the model did not see before, in other words, unseen data instances). Note that the authors did not monitor the validation loss or accuracy but tested the model on the unseen data after the training. Additionally, the success of the model on the unseen data indicates if the model is overfitted to the training data and can be generalized to other datasets.

The test results are evaluated by using one regression and two classification metrics. The reason why a regression metric is used on a classification problem is to measure the error of faulty predictions. In a simple vibration-based damage detection problem in SHM, a prediction score such as 0.77 can be both interpreted as damaged data (with a threshold assumption of 0.5 and converted into the label of "1" which indicates damage data) and the quantification of the damage data (such as loosening a bolt not 100% but with a percentage of 77). Therefore, it is prominent to distinguish between two indications. The Mean Absolute Error (MAE) is used as the regression metric:

$$MAE = \frac{\sum_{i=1}^{n}|y_i - x_i|}{n} \qquad (4)$$

Where $n$ is the total number of samples, $i$ is the index of the sample, $y$ is the predicted value, and $x$ is the actual value of the sample. MAE metric is also used for many regression and classification tasks in both ML and DL studies. For the classification metrics, Classification Accuracy (CA) and Average Precision (AP) scores are utilized. The CA is one of the most used metrics in ML and DL studies that simply measures the total correct predictions over total predictions:

$$Classification\ Accuracy = \frac{Number\ of\ correct\ predictions}{Total\ number\ of\ predictions} \qquad (5)$$

In order to use the CA, a threshold has to be assigned in the domain of prediction scores to convert the prediction score into the closest label (in this study the used labels are "0" undamaged and "1" damaged). This study used 0.50 as the threshold. The prediction scores made at 0.50 and above are converted to 1 and the others are to 0. Yet, evaluating a model based on one threshold value can be misleading about the performance of the model. The AP is one of the most used metrics which gives average precision at all possible thresholds, especially employed for benchmarking different DL models on various datasets. It is a very useful metric to compare the successes of different models. The AP combines and summarizes the precision and recall curve into one value which represents the weighted summation of precisions at the different thresholds. The weight is defined as the increase in recall from each succeeding threshold. The precision is the ratio of true positive over the sum of true positive and false positive. This metric implies the frequency of correct predictions at every prediction; thus, it reflects how reliable the model is in predicting the samples as positive. The recall is the ratio of true positive over the sum of true positive and false negative which implies the model's ability to classify positive samples and only interest in how the positive samples are classified. Briefly, the AP is the area under the precision-recall curve. An area of 1.0 means that the classifier is a perfect model and 0.5 means the classifier is a poor model. The formulation of AP can be shown as:

$$AP = \sum_n [Recall(n) - Recall(n-1)] \times Precision(n) \qquad (6)$$

After $M_2$ is used on the test data to predict the damaged tensors for each scenario in Fig. 2, the prediction results for the tested tensors and their corresponding ground truths (actual labels) are bar plotted in Fig. 12. In other words, the prediction score of $M_2$ for each tensor is shown as the blue bar and the corresponding ground truth is shown as the orange bar. Also, the MAE, CA, and AP metrics are computed and displayed on the top right corner of the bar plots for each scenario in Figs. 12-17. At first glance, one can catch the close prediction scores to ground truths in these figures. Yet with a closer look, there are a few incorrect prediction results that are circled with red. Starting with Scenario#0, the regression and classification metrics show that the model predicts 100% CA, 1.00 AP, and $4.7 \times 10^{-3}$ MAE. That means the model yields consistent results on every tensor with no inaccurate classification. Note that Scenario#0 (no synthetic tensor is involved) was created as a "reference" scenario for benchmarking purposes. For the rest of the scenarios, from Scenario#1 to Scenario#5, the inclusion ratio of the number of synthetic tensors in the damage test dataset is gradually decreased. This change slightly impacted the classification results. Firstly, the corresponding scenarios in Figs. 13-17 includes one incorrect prediction on different tensors that are circled in red color. While the $M_2$ is more confident in its predictions in Scenario#1-2-3, for Scenario#4 and Scenario#5 it remains uncertain on a few tensors. The incorrect predictions caused the MAE metric to increase about 6.5 times from $4.7 \times 10^{-3}$ to $0.3 \times 10^{-1}$ ranges when comparing Scenario#0 to other scenarios. This indicates that the predictions of $M_2$ include more errors for the other scenarios than the reference scenario. Having one incorrect prediction caused the CA score to decrease from 100% to 97%, and also the AP score to decrease from 1.00 to 0.97 for Scenario#1 and Scenario#2. The change in AP score is basically the result of the high confidence of the model on the data for Scenario#1 and Scenario#2. Since the model is very sure of its predictions, the AP score always contains one incorrect prediction at every threshold value unless the threshold is selected very close to 0 or 1. While the inclusion of synthetic tensors slightly changed the MAE metric, the CA and AP metrics experienced a negligible amount of decrease which may be critical for vibration-based damage detection. The classification metric is more used for level-1 damage diagnostics (damage detection) and the error metrics can be more beneficial for level-2 damage diagnostics (damage quantification) since the damage quantification is carried out based on the resulted error. It is of the opinion that the model is overconfident about the test data which caused the prediction result on test samples wrong (maximum 1 incorrect prediction in each scenario). Taking into account one incorrect prediction out of 30 predictions for Scenario#1-5, the prediction results on the test dataset can be concluded as satisfactory for this dataset. It is, however, critical to evaluate the extent of incorrect predictions to be accepted considering the overall structural safety as a future study.

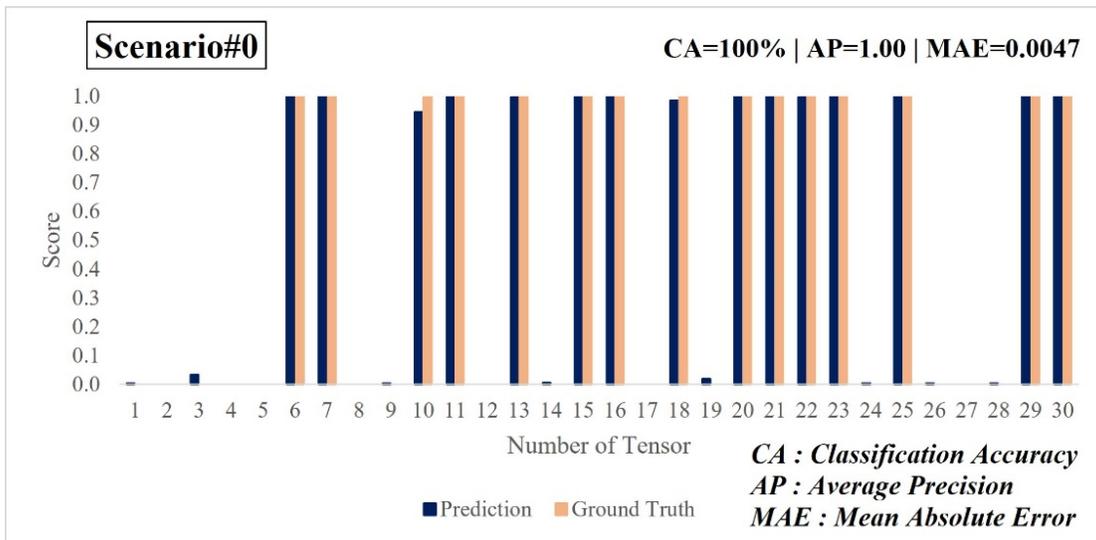

**Fig. 13.** Testing results of $M_2$ for Scenario#0

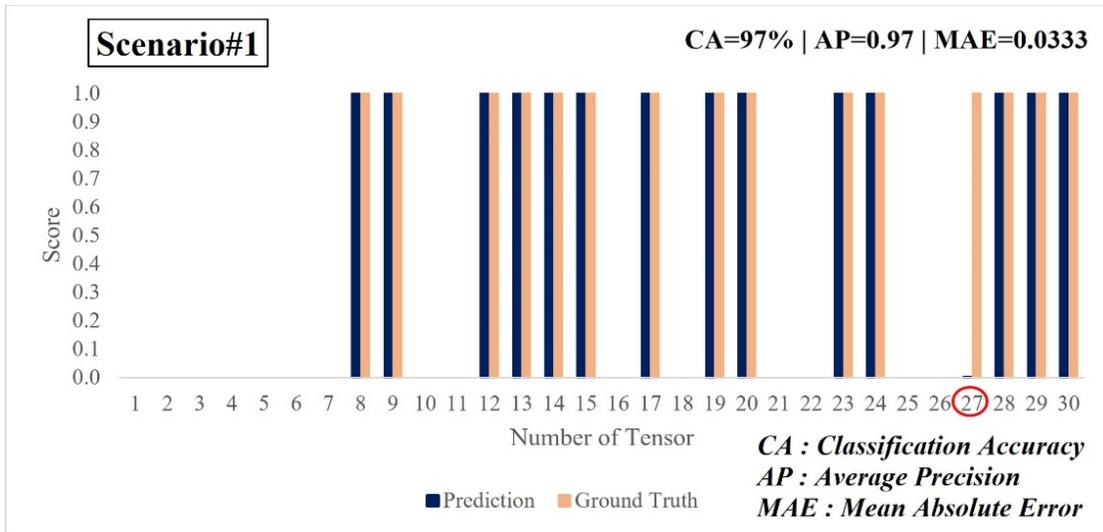

**Fig. 14.** Testing results of $M_2$ for Scenario#1

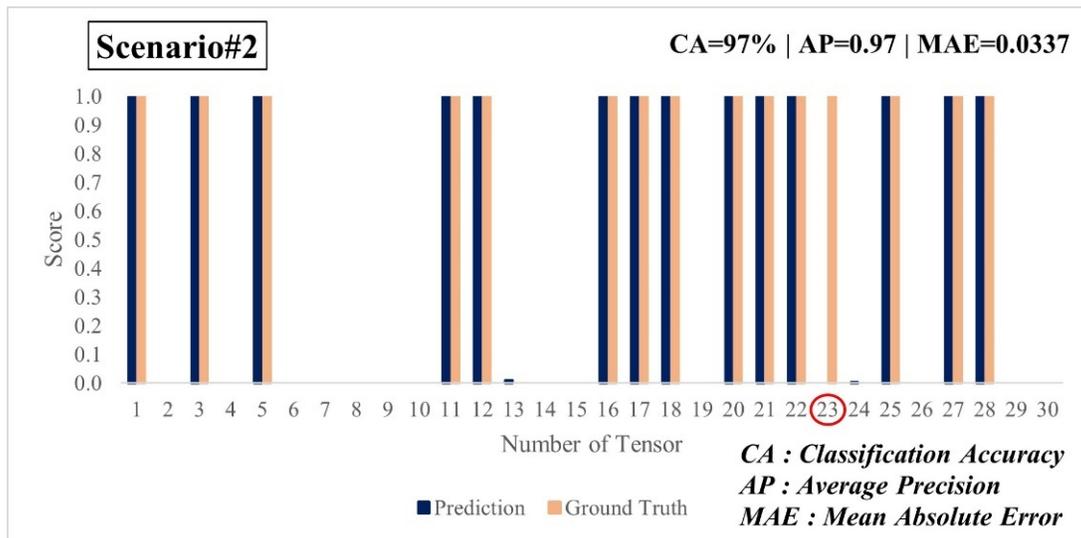

**Fig. 15.** Testing results of $M_2$ for Scenario#2

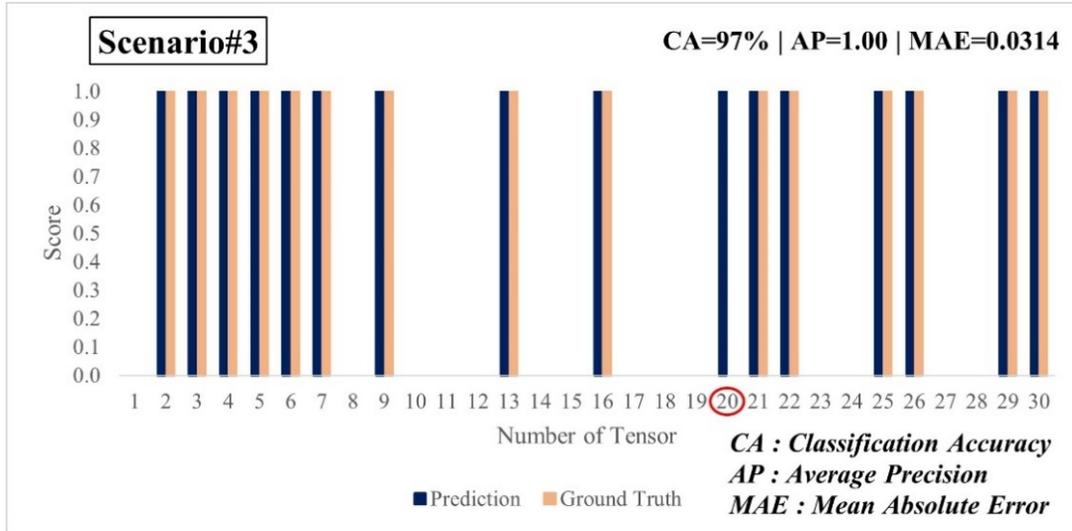

**Fig. 16.** Testing results of $M_2$ for Scenario#3

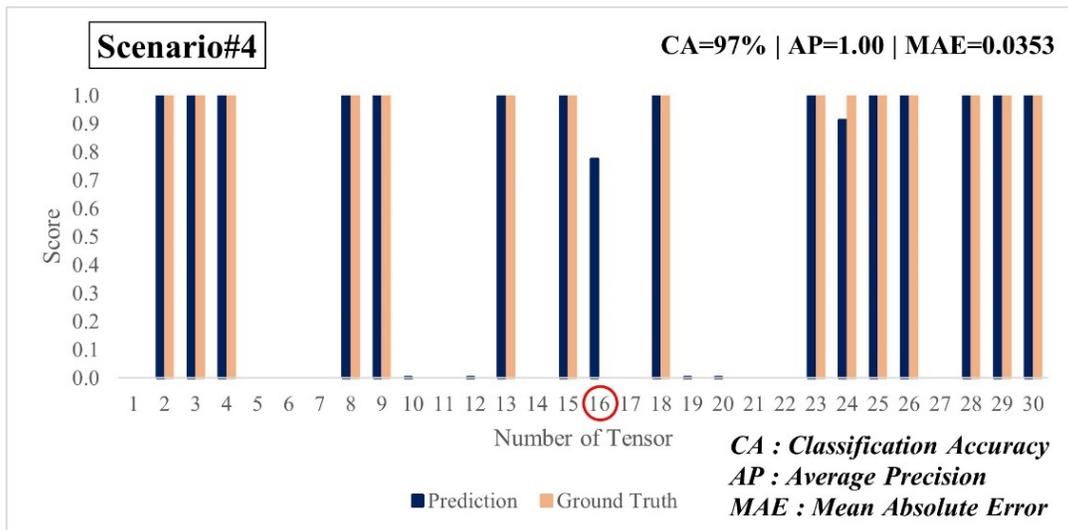

**Fig. 17.** Testing results of $M_2$ for Scenario#4

**Fig. 17.** Testing results of $M_2$ for Scenario#4

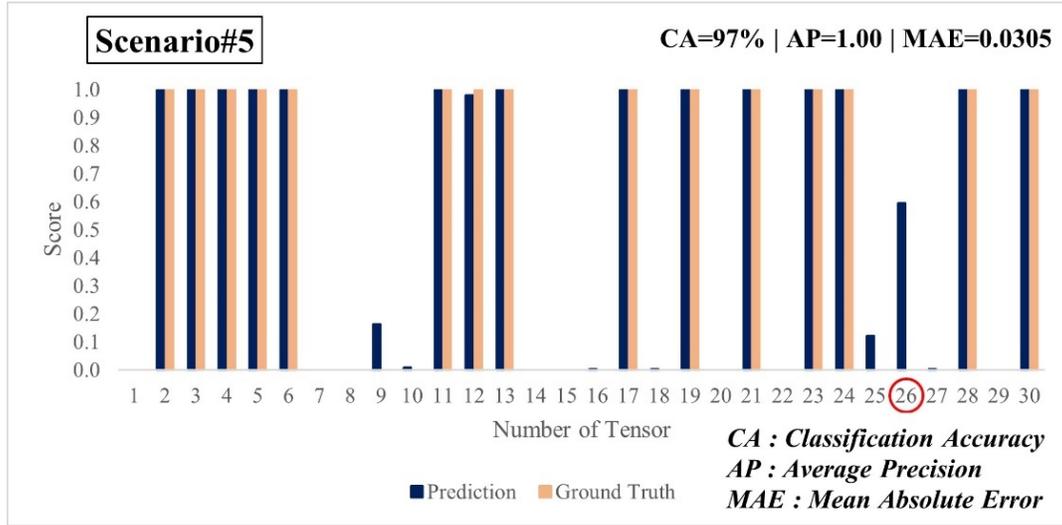

**Fig. 18.** Testing results of $M_2$ for Scenario#5

**5) Summary and conclusions**

To establish an ML or DL model for damage detection purposes, a substantial amount of data is necessary for the model training procedure. Yet, this can be very expensive and time-consuming since obtaining dynamic response data from civil structures is a very difficult and expensive task which creates the data scarcity problem in SHM, and this leads to a class imbalance problem (such as having more undamaged class than damaged class). The data scarcity problem hinders the use of AI methods in SHM applications. Recently, the researchers took advantage of AI methods such as ML and DL algorithms for vibration-based structural damage detection and they produced excellent results. Particularly the DL methods perform exceptionally well, yet it requires a large amount of data to operate. This study utilized One Dimensional Wasserstein Deep Convolutional Generative Adversarial Networks using Gradient Penalty (1-D WDCGAN-GP) to generate synthetic damaged acceleration data and validated the generated datasets with quantitative and qualitative methods. Then, five different scenarios are created to reflect the real-life scenarios that may be faced in the SHM for the purpose of vibration-based damage detection. Each of these scenarios is augmented with different levels (ratios) of synthetic vibration tensors in the training dataset of a damage detection model, 1-D DCNN (One Dimensional Deep Convolutional Neural Network). Then, for augmentation of each scenario (Scenario 1, 2, 3, 4, 5), the 1-D DCNN model is used to perform damage detection on the raw acceleration tensors (nonparametric damage detection). The prediction results of 1-D DCNN are evaluated with regression and classification metrics. While the prediction scores for the synthetically augmented dataset scenarios yielded 97% classification accuracy, the real dataset yielded 100% classification accuracy. In other words, when the 1-D DCNN is trained with 1-D WDCGAN-GP augmented damaged class and real undamaged class, its prediction capability falls in 3% error margin compared to the benchmark case (Scenario 0, real undamaged and real damaged classes). Thus, from the perspective of the 1-D DCNN model, the augmented damaged class is almost indistinguishable from the real damaged datasets. The main conclusions of this study can be listed as the following:

- 1-D WDCGAN-GP can be used to augment the imbalanced dataset, which is caused by the data scarcity problem in SHM of civil structures, a well-known problem in the field. Imbalanced data classes can be detrimental to ML and DL-based structural damage detection applications. By generating synthetic datasets with 1-D WDCGAN-GP, the training datasets can be balanced. Thus, this increases the efficiency of damage detection applications and possibly reduces the need for data collection from civil structures.

- Additional research is needed for level-1 damage diagnostics (detection) and level-2 damage diagnostics (quantification) using 1-D WDCGAN-GP. New studies should investigate the effect of the degree of data augmentation with 1-D WDCGAN-GP on the damage prediction results. Also, further developments and investigations are necessary for evaluating the GAN-generated vibration signals (1-D) since the existing evaluation methodologies consider image (2-D) data.

- The methodology presented in this paper is needed to be tested on real civil structures for further validation to be used in practice. Lastly, the effect of the given noise in 1-D WDCGAN-GP should be investigated and evaluated based on the generated data samples.

**Data availability statement**

Vibration data used in this study was made available in Abdeljaber et al., 2017 and was published in a benchmark format in Avci et al. (2022). Some or all used models, codes, and detailed results are available from the corresponding author of this paper upon request.

**Acknowledgement**

The authors would like to thank members of the CITRS (Civil Infrastructure Technologies for Resilience and Safety) Research Initiative at the University of Central Florida. The second author would like to acknowledge the support of the National Aeronautics and Space Administration (NASA) Award No. 80NSSC20K0326 and by the U.S. National Science Foundation (NSF) Division of Civil, Mechanical and Manufacturing Innovation (grant number 1463493). The third author would like to acknowledge the support and guidance of Dr Serkan Kiranyaz.

**Nomenclature**

The following symbols are used in this paper:

| Symbol | Description |
|---|---|
| $[a_u{}^{t256}]$ | Undamaged acceleration data for 256 seconds |
| $[a_d{}^{t256}]$ | Damaged acceleration data for 256 seconds |
| $n[a_u{}^{t1}]$ | $n$ amount of undamaged acceleration data for 1 second |
| $n[a_d{}^{t1}]$ | $n$ amount of damaged acceleration data for 1 second |
| $n[a_{d,s}{}^{t1}]$ | $n$ amount of synthetic damaged acceleration data for 1 second |
| $M_1$ | Used 1-D WDCGAN-GP model in the paper |
| $M_2$ | Used 1-D DCNN model in the paper |